\documentclass[11pt]{article}

\usepackage[a4paper,
            top=1in,
            bottom=1in,
            left=1in,
            right=1in]{geometry}

\usepackage{times}
\usepackage{microtype}

\usepackage{amsmath,amsfonts,bm}

\def\eqref#1{equation~\ref{#1}}

\def\1{\bm{1}}

\DeclareMathAlphabet{\mathsfit}{\encodingdefault}{\sfdefault}{m}{sl}
\SetMathAlphabet{\mathsfit}{bold}{\encodingdefault}{\sfdefault}{bx}{n}

\usepackage{amsmath}
\usepackage{amssymb}

\usepackage{algorithm}
\usepackage{algorithmicx}
\usepackage{algcompatible}
\usepackage{algpseudocode}

\usepackage{booktabs}
\usepackage{longtable}
\usepackage{multirow}
\usepackage{siunitx}
\usepackage[table]{xcolor}

\usepackage{graphicx}
\usepackage{placeins}

\usepackage{hyperref}
\usepackage{url}

\title{
    \textbf{MedHarnessBench: A Unified Benchmark for Healthcare Agent Evaluation}
}

\author{
    Rui Tang$^{1}$
    \qquad
    Hongqiu Wang$^{2}$ \\[0.5em]
    $^{1}$The University of Hong Kong \\
    $^{2}$The Hong Kong University of Science and Technology (Guangzhou)
}

\date{}

\newcommand{\protocol}{\textsc{Medical Harness Protocol}}
\newcommand{\benchmark}{\textsc{MedHarnessBench}}

\definecolor{tablegray}{gray}{0.9}
\definecolor{tablehead}{HTML}{F1F3F5}
\definecolor{benchphys}{HTML}{F2EEF8}
\definecolor{benchadmin}{HTML}{EEF3F8}
\definecolor{benchcta}{HTML}{EEF8F2}
\definecolor{bestcell}{HTML}{F4E7BF}

\newcommand{\best}[1]{%
    \cellcolor{bestcell}\bfseries #1%
}

\newcommand{\gain}[1]{%
    \textcolor{green!55!black}{%
        \scriptsize\(\mathbf{(+#1)}\)%
    }%
}

\newcommand{\loss}[1]{%
    \textcolor{red!75!black}{%
        \scriptsize\(\mathbf{(-#1)}\)%
    }%
}

\newcommand{\dcell}[2]{%
    \multicolumn{1}{c}{#1\,#2}%
}

\begin{document}

\maketitle

\begin{abstract}
Healthcare agents must transform longitudinal clinical evidence into reliable actions through heterogeneous tools and stateful workflows, conditions under which model-centric systems may lose context, misuse tools, fail to verify consequential transitions, or cross authorization boundaries. Existing medical-agent benchmarks primarily emphasize terminal task outcomes, making it difficult to determine whether execution failures arise from the agent model, the runtime harness, or their interaction. We present the \textbf{\textit{Medical Harness Protocol}}, an outcome-preserving framework that independently evaluates seven healthcare-specific process dimensions: \textit{Care Action Fulfillment}, \textit{Clinical Resource Competence}, \textit{Patient--Evidence Alignment}, \textit{Clinical Workflow Continuity}, \textit{Clinical Auditability}, \textit{Closed-Loop Verification}, and \textit{Clinical Conduct Boundaries}. A normalized intervention trace further tracks whether runtime support is available, activated, delivered, adopted by the agent, and propagated into subsequent behavior. We instantiate the protocol as \textbf{\textit{MedHarnessBench}}, comprising 342 executable tasks across FHIR/EHR operations, healthcare-administration interfaces, and multimodal clinical tools. Across more than 16,000 controlled task executions, MedHarness yields broad improvements in terminal task success, intermediate subtask completion, and healthcare-specific process quality. We further evaluate MedHarness against multiple general-purpose harnesses and use controlled single-fault perturbations to verify the sensitivity and discrimination of the seven-dimensional evaluation. These results demonstrate that healthcare-specific runtime support can substantially improve both task completion and execution quality, while revealing failures that terminal outcomes alone cannot capture. Code and resources are available at \url{https://github.com/Tomry1114/Awesome-Agent-Harness-for-Healthcare}.
\end{abstract}

\section{Introduction}
\label{sec:introduction}

Language-model agents increasingly operate through interactive workflows that retrieve evidence, invoke tools, observe feedback, and execute multi-step actions~\cite{yao2023react,shinn2023reflexion}. In healthcare, these workflows span longitudinal patient records, medical images, clinical messages, administrative interfaces, and heterogeneous medical tools~\cite{jiang2025medagentbench,schmidgall2024agentclinic}. Reliable execution therefore requires appropriate evidence and resource use, continuous workflow state, verification of consequential actions, and compliance with clinical and authorization boundaries~\cite{arora2025healthbench,liu2026harnessaudit}. Yet a plausible response may still rely on stale evidence, an unexecuted order, an unverified write, or an unauthorized action, while locally correct execution may remain unrecognized when the final task fails.

Recent medical-agent benchmarks extend evaluation from static answers to executable clinical, administrative, and multimodal workflows~\cite{liu2026physicianbench,bedi2026healthadminbench,ashraf2026medcta}. However, terminal outcomes and environment-specific checkpoints provide limited evidence about the runtime layer that constructs context, exposes tools, maintains state, supports recovery, verifies actions, and enforces permissions. Consequently, failures caused by unavailable resources, invalid tool use, missing verification, or incorrect model decisions may appear indistinguishable, while unreliable intermediate behavior may remain hidden behind successful task completion.

We formalize this runtime layer as the \textit{Medical Harness} and introduce the \textit{Medical Harness Protocol} for evaluating its contribution to healthcare-agent execution. Building on the ETCLOVG taxonomy of Li et al.~\cite{li2026agentharness}, we specialize Execution, Tooling, Context, Lifecycle, Observability, Verification, and Governance as Care Action Fulfillment, Clinical Resource Competence, Patient--Evidence Alignment, Clinical Workflow Continuity, Clinical Auditability, Closed-Loop Verification, and Clinical Conduct Boundaries. The protocol retains each environment's native task metrics while evaluating these seven healthcare-specific process dimensions separately.

We instantiate the protocol as \textit{MedHarnessBench}, comprising 342 executable tasks from PhysicianBench, HealthAdminBench, and MedCTA across FHIR/EHR operations, healthcare-administration interfaces, and multimodal clinical tools~\cite{liu2026physicianbench,bedi2026healthadminbench,ashraf2026medcta}. Across eight agents, MedHarness improves Pass@3 in 19 of 24 agent--environment configurations and Subtask performance in 22 of 24. Comparisons with general-purpose harnesses demonstrate the value of healthcare-specific runtime support, while controlled single-fault perturbations verify the sensitivity and discrimination of the seven process dimensions.

Our contributions are:
\begin{itemize}
    \item \textbf{Medical Harness.}
    We formalize the healthcare-specific runtime that coordinates clinical evidence, heterogeneous tools, workflow state, recovery, verification, and authorization around an AI agent.

    \item \textbf{Medical Harness Protocol.}
    We introduce seven healthcare-specific process dimensions for evaluating runtime execution separately from benchmark-native task performance.

    \item \textbf{MedHarnessBench.}
    We evaluate eight agents on 342 executable tasks through controlled Native--MedHarness comparisons, general-purpose harness baselines, and single-fault perturbations.
\end{itemize}
\section{Related Work}
\label{sec:related}

\subsection{Healthcare Agent Benchmarks}

Medical-agent evaluation has expanded from static question answering to interactive and executable healthcare workflows. AgentClinic simulates multimodal clinical diagnosis~\cite{schmidgall2024agentclinic}, while HealthBench evaluates open-ended healthcare conversations using physician-authored rubrics~\cite{arora2025healthbench}. More recent benchmarks introduce operational environments: MedAgentBench provides a FHIR-compatible EHR setting~\cite{jiang2025medagentbench}; PhysicianBench evaluates 100 long-horizon physician tasks~\cite{liu2026physicianbench}; HealthAdminBench covers 135 GUI-based administrative tasks~\cite{bedi2026healthadminbench}; and MedCTA evaluates clinically validated multimodal tool use~\cite{ashraf2026medcta}. HealthAgentBench broadens evaluation across agentic healthcare settings~\cite{liu2026healthagentbench}, and $\chi$-Bench introduces policy-rich, multi-role workflows~\cite{chen2026chibench}.

These benchmarks substantially improve task realism but primarily evaluate complete agent configurations through final answers, environment states, rubrics, or task-specific checkpoints. Such measurements provide limited attribution of failures to model decisions, context construction, tool mediation, workflow-state management, verification, or authorization. MedHarnessBench addresses this gap by retaining benchmark-native task evaluation while separately measuring the healthcare-specific runtime processes surrounding the model.

\subsection{Agent Harnesses and Runtime Evaluation}

Language-model agents commonly operate through runtime loops that connect reasoning, actions, tools, and environmental observations~\cite{yao2023react}. Reflection, memory, and reusable skills can alter agent behavior without changing the underlying model~\cite{shinn2023reflexion,wang2023voyager,zhao2023expel}, while SWE-agent demonstrates the importance of designing interaction interfaces around model capabilities~\cite{yang2024sweagent}. Together, these studies establish the runtime as a substantive component of agent performance.

Recent work formalizes this runtime as an agent harness. Natural-Language Agent Harnesses represent runtime policies through editable specifications and execution contracts~\cite{pan2026natural}; Harness-Bench evaluates model--harness configurations under shared tasks and environments~\cite{yao2026harnessbench}; and HarnessAudit examines execution and permission violations hidden behind benign outputs~\cite{liu2026harnessaudit}. Harness-aware evaluation further treats runtime configuration as an explicit experimental condition~\cite{zhang2026disclosing}, while Li et al.~\cite{li2026agentharness} organize harness functionality through the ETCLOVG taxonomy. These formulations are primarily designed for coding, computer-use, and general-purpose agents and do not explicitly address longitudinal patient evidence, heterogeneous medical tools, persistent clinical writes, closed-loop verification, or role-dependent authorization. The Medical Harness specializes this runtime abstraction for healthcare and makes its contribution measurable separately from the underlying agent model.

\section{Method}
\label{sec:method}

\begin{figure*}[t]
    \centering
    \includegraphics[width=\textwidth]{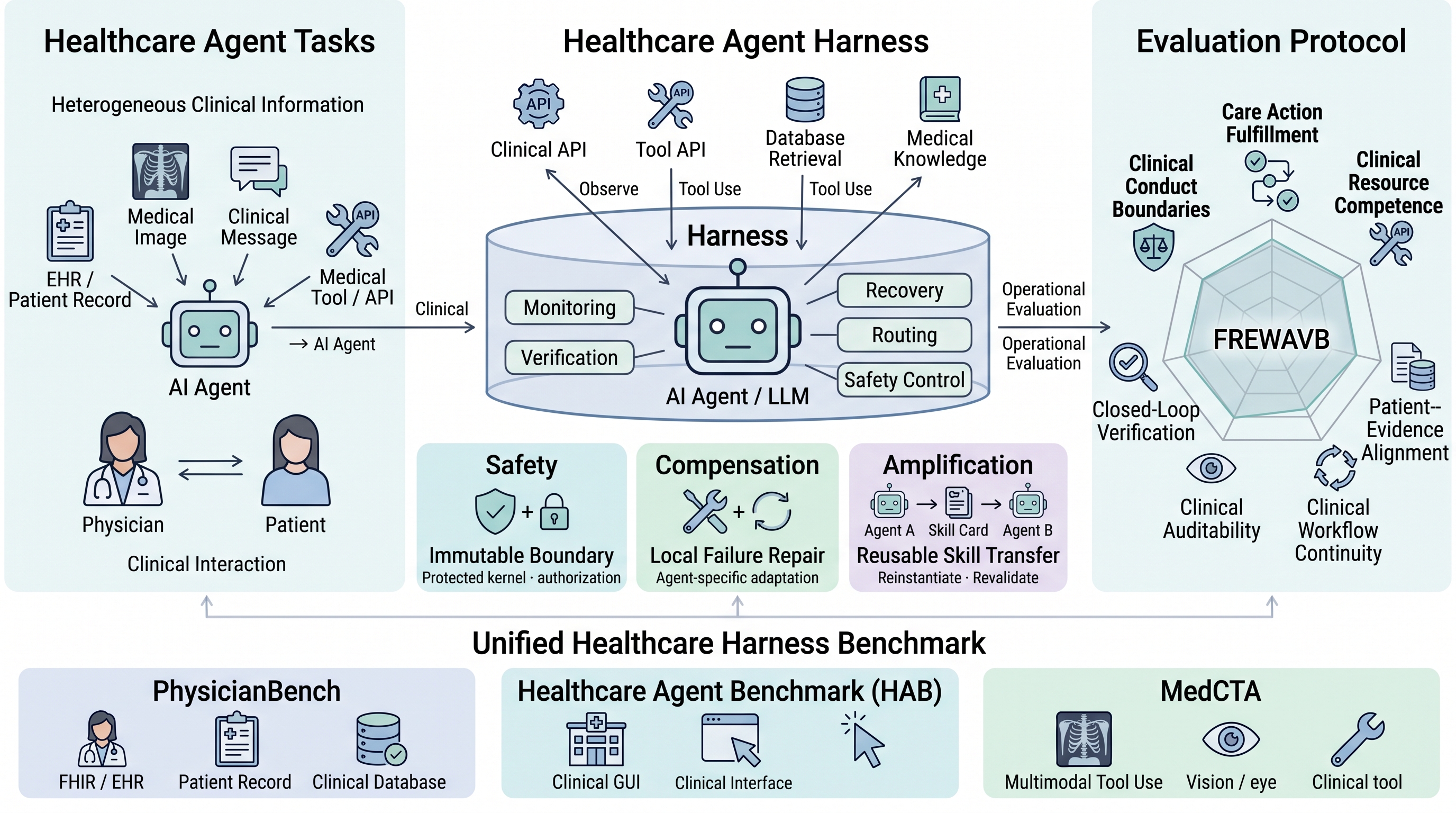}
    \vspace{-2mm}
    \caption{
    Overview of MedHarnessBench. Healthcare agents operate over heterogeneous evidence and interfaces, while MedHarness mediates their interactions with clinical APIs, general tools, databases, and medical knowledge through monitoring, verification, recovery, routing, and safety control. The benchmark evaluates harness-mediated execution through seven healthcare-specific process dimensions: Care Action Fulfillment, Clinical Resource Competence, Patient--Evidence Alignment, Clinical Workflow Continuity, Clinical Auditability, Closed-Loop Verification, and Clinical Conduct Boundaries. It integrates 342 executable tasks from PhysicianBench, HealthAdminBench, and MedCTA, spanning FHIR/EHR operations, clinical GUI workflows, and multimodal medical tool use.
    }
    \vspace{-3mm}
    \label{fig:medharnessbench_overview}
\end{figure*}

\subsection{Problem Formulation}
\label{sec:problem}

Healthcare-agent performance emerges from the interaction between a model and the runtime that supplies evidence, tools, state, verification, and authorization. We isolate this runtime contribution by treating the agent, harness configuration, executable environment, and task as distinct components.

Let \(A_i\) denote a healthcare agent, \(\mathcal{E}\) an executable healthcare environment, and \(x\sim\mathcal{D}_{\mathcal E}\) a task. Each agent is evaluated under two controlled runtime configurations. The \emph{Native} configuration \(H_{\mathrm{nat}}\) retains the agent's original execution interface, whereas the \emph{MedHarness} configuration \(H_{\mathrm{med}}\) augments that interface with healthcare-specific evidence acquisition, tool mediation, workflow-state tracking, monitoring, recovery, verification, and authorization control.

A task execution under configuration \(c\) is represented as
\begin{equation}
\tau_i^{c}
\sim
\operatorname{Run}(A_i,H_c,\mathcal{E},x;\xi),
\qquad
c\in\{\mathrm{nat},\mathrm{med}\},
\label{eq:run}
\end{equation}
where \(\xi\) captures model and environment stochasticity. Each trajectory records model outputs, tool or GUI actions, observations, harness interventions, state transitions, evaluation checkpoints, and persistent artifacts.

Within each Native--MedHarness pair, we hold the task, agent, system prompt, tool inventory, transport configuration, initial environment state, and execution budget fixed, varying only the healthcare-specific runtime support. Each run retains the benchmark-native Outcome, is evaluated along seven process dimensions, and records whether MedHarness interventions activate, reach the agent, alter its behavior, and propagate through the workflow. Figure~\ref{fig:medharnessbench_overview} summarizes this controlled evaluation.

\subsection{MedHarnessBench}
\label{sec:benchmark}

Existing healthcare-agent benchmarks provide environment-specific Outcome metrics for task completion, but expose limited evidence about the runtime process that produced them. MedHarnessBench preserves these native endpoints and adds a shared process-level protocol over execution trajectories. The resulting evaluation separates what the agent achieves from how the agent--harness configuration achieves it, enabling process diagnosis without replacing the original task objective.

\subsubsection{Benchmark Environments}
\label{sec:environments}

MedHarnessBench comprises 342 tasks from PhysicianBench, HealthAdminBench, and MedCTA while preserving each benchmark's native tasks, tools, interfaces, and Outcome evaluator. PhysicianBench executes FHIR/EHR operations against a HAPI-FHIR R4 server restored before each task; HealthAdminBench evaluates healthcare-administration workflows through a Playwright-controlled clinical portal; and MedCTA provides a multimodal sandbox with image analysis, OCR, retrieval, and calculation tools. Table~\ref{tab:benchmark} summarizes their execution settings and harness demands.

\begin{table*}[t]
\caption{Executable healthcare environments in MedHarnessBench.}
\label{tab:benchmark}
\centering
\scriptsize
\setlength{\tabcolsep}{2.5pt}
\renewcommand{\arraystretch}{1.08}
\begin{tabular}{
@{}
p{0.16\textwidth}
p{0.06\textwidth}
p{0.31\textwidth}
p{0.41\textwidth}
@{}
}
\toprule
\textbf{Benchmark}
& \textbf{\#Tasks}
& \textbf{Execution setting}
& \textbf{Harness demands} \\
\midrule

PhysicianBench
& 100
& FHIR API; clinical retrieval, ordering, and documentation
& Evidence grounding; structured writes; authorization; write verification \\

HealthAdminBench
& 135
& Clinical GUI; prior authorization, appeals, and submission
& State tracking; field completion; recovery; submission verification \\

MedCTA
& 107
& Multimodal tools; image and report analysis, retrieval, and calculation
& Tool selection; argument validity; evidence integration; result use \\

\midrule
\textbf{Total}
& \textbf{342}
& Three executable healthcare settings
& Process control across API, GUI, and multimodal workflows \\

\bottomrule
\end{tabular}
\end{table*}

A substrate-specific adapter maps these heterogeneous interactions into a shared semantic trace without altering their native execution semantics. The trace records action roles, execution status, workflow progress, state changes, evidence provenance, verification events, failure attribution, and terminal state. Model generation, environment execution, and semantic evaluation are kept separate to distinguish proposed actions, executed effects, and evaluator judgments.

\subsubsection{Healthcare Process Evaluation}
\label{sec:protocol}

Native Outcome determines whether a healthcare task succeeds but does not identify the runtime conditions that produced the result. We therefore complement it with process-level measurements of the agent--harness trajectory. Building on the ETCLOVG taxonomy of Li et al.~\cite{li2026agentharness}, we specialize its seven layers for healthcare workflows. Four Efficiency dimensions characterize action completion, resource use, evidence alignment, and workflow continuity, while three Safety dimensions characterize auditability, verification, and clinical conduct. Table~\ref{tab:healthcare-process} summarizes their operational definitions.

\begin{table*}[t]
\caption{Healthcare-specific process dimensions in MedHarnessBench.}
\label{tab:healthcare-process}
\centering
\scriptsize
\setlength{\tabcolsep}{3pt}
\renewcommand{\arraystretch}{1.08}
\begin{tabular}{
@{}
p{0.11\textwidth}
p{0.27\textwidth}
p{0.56\textwidth}
@{}
}
\toprule
\textbf{Category} & \textbf{Dimension} & \textbf{Operational focus} \\
\midrule

\multirow{4}{*}{\textbf{Efficiency}}
& Care Action Fulfillment
& Completion of required care actions to their intended state. \\

& Clinical Resource Competence
& Authorized, semantically valid, and productive use of clinical resources. \\

& Patient--Evidence Alignment
& Acquisition of required evidence and binding to the intended clinical subject. \\

& Clinical Workflow Continuity
& Ordered workflow progression, recovery from failed obligations, and appropriate termination. \\

\midrule

\multirow{3}{*}{\textbf{Safety}}
& Clinical Auditability
& Reconstructability of execution state, failures, recovery, and reliability. \\

& Closed-Loop Verification
& Confirmation of decision-critical preconditions and post-action state changes. \\

& Clinical Conduct Boundaries
& Compliance with protected information, subject, provenance, scope, risk, and reporting boundaries. \\

\bottomrule
\end{tabular}
\end{table*}

\paragraph{Healthcare Process Operationalization.}
We distinguish between indicator selection, which determines whether an indicator directly establishes a healthcare process dimension, and trajectory-level applicability, which determines whether that indicator can be evaluated in a particular execution. Let \(\mathcal{R}_d\) denote the candidate indicators for dimension \(d\). The selected indicator set is
\begin{equation}
\mathcal{K}_d
=
\{r\in\mathcal{R}_d:
\textsc{Establishes}(r,d)
\wedge
\textsc{Validated}(r)\}.
\end{equation}
Here, \textsc{Establishes} requires direct evidence for \(d\), while \textsc{Validated} excludes indicators that introduce counter-incentives or systematically favor weaker execution. Selection does not require correlation with Outcome because correct process behavior and terminal task success may diverge.

For a realized trajectory \(\tau\), only selected indicators with a valid measurement opportunity are activated:
\begin{equation}
\mathcal{A}_d(\tau)
=
\{r\in\mathcal{K}_d:
\textsc{Opportunity}_r(\tau)\}.
\end{equation}
The corresponding dimension score is
\begin{equation}
S_d(\tau)=
\begin{cases}
g_d(\tau,\mathcal{A}_d(\tau)),
& \textsc{EvidenceOK}_d(\tau),\\
\bot,
& \text{otherwise}.
\end{cases}
\end{equation}
Here, \textsc{EvidenceOK} requires non-empty applicable evidence and dimension-specific coverage. Clinical Workflow Continuity additionally requires sufficient core workflow evidence, while Clinical Conduct Boundaries requires an available independent evaluator. Unsupported trajectories receive \(\bot\), and aggregate scores are reported together with their applicable support.

\paragraph{Clinical Workflow Efficiency Evaluation.}
The four Efficiency dimensions evaluate whether an agent completes required care actions, uses clinical resources competently, aligns evidence with the intended patient, and maintains continuity throughout the workflow.

Care Action Fulfillment combines completion of the best valid required-operation path, \(C_{\mathrm{op}}\), with emitted-action validity, \(M_{\mathrm{act}}\):
\begin{equation}
g_{\mathrm{care}}
=
\begin{cases}
C_{\mathrm{op}}
[\lambda+(1-\lambda)M_{\mathrm{act}}],
& C_{\mathrm{op}}\text{ applicable},\\
M_{\mathrm{act}},
& \text{otherwise}.
\end{cases}
\end{equation}
This formulation makes completion the dominant signal: valid actions can qualify achieved completion but cannot create credit for unfinished work.

Clinical Resource Competence is macro-averaged across selected resource indicators:
\begin{equation}
g_{\mathrm{resource}}
=
\frac{1}{|\mathcal{A}_{\mathrm{resource}}|}
\sum_{r\in\mathcal{A}_{\mathrm{resource}}}
\frac{1}{|\mathcal{I}_r|}
\sum_{i\in\mathcal{I}_r}s_r(i),
\end{equation}
where \(\mathcal{I}_r\) contains the invocation events exposing indicator \(r\). Averaging first within each indicator prevents frequently exercised resources from dominating the score.

Patient--Evidence Alignment combines required-evidence recall with subject correctness:
\begin{equation}
g_{\mathrm{align}}
=
w_R
\frac{|\mathcal M(\tau)|}{|\mathcal U_{\mathrm{req}}(x)|}
+
w_M
\mathbb{1}[\sigma(\tau)=\sigma^\star(x)],
\end{equation}
where \(\mathcal{M}(\tau)\) is the retrieved subset of required evidence, \(\mathcal{U}_{\mathrm{req}}(x)\) is the required evidence set, and \(\sigma^\star(x)\) is the intended clinical subject. The weights satisfy \(w_R+w_M=1\). This distinction prevents relevant evidence from receiving full credit when it is bound to the wrong patient or clinical context.

Clinical Workflow Continuity evaluates ordering only over constraints activated by the realized trajectory:
\begin{equation}
s_{\mathrm{order}}
=
\frac{\sum_{c\in\mathcal P(\tau)}
\nu_c\mathbb{1}[\mathrm{pred}_c\prec\mathrm{succ}_c]}
{\sum_{c\in\mathcal P(\tau)}\nu_c},
\end{equation}
where \(\mathcal{P}(\tau)\) contains constraints whose successor events are observed and \(\nu_c\) is the importance weight of constraint \(c\). Recovery, stagnation, readiness, and termination are evaluated under the same opportunity-based principle.

\paragraph{Clinical Workflow Safety Evaluation.}
The three Safety dimensions evaluate whether healthcare-agent execution remains auditable, verifies consequential actions in closed loop, and respects protected clinical boundaries.

Clinical Auditability macro-averages reconstructability evidence across capability families:
\begin{equation}
g_{\mathrm{audit}}
=
\frac1{|\mathcal Q|}
\sum_{q\in\mathcal Q}
\frac1{|\mathcal F_q|}
\sum_{f\in\mathcal F_q}
\bar s_{q,f},
\end{equation}
where \(\mathcal{Q}\) denotes the applicable capability families and \(\mathcal{F}_q\) contains their selected evidence fields. Delivery fidelity and error transparency are retained as diagnostic signals unless they directly establish reconstructability.

Closed-Loop Verification aggregates applicable verification evidence using trajectory-dependent weights:
\begin{equation}
g_{\mathrm{verify}}
=
\frac{
\sum_{r\in\mathcal A_{\mathrm{verify}}}
w_r^{(\tau)}s_r}
{\sum_{r\in\mathcal A_{\mathrm{verify}}}w_r^{(\tau)}}.
\end{equation}
The weights distinguish stronger adjudicated evidence from automatically detectable confirmation. Reading evidence or invoking a checking mechanism alone does not establish that a consequential action was verified.

Clinical Conduct Boundaries enforce protected-information, subject, provenance, scope, risk, and reporting constraints:
\begin{equation}
g_{\mathrm{conduct}}
=
\begin{cases}
0,&\textsc{CriticalViolation},\\
\bot,&\text{independent evaluator unavailable},\\
\omega G_{\mathrm{rules}}+(1-\omega)B_{\mathrm{subj}},
&B_{\mathrm{subj}}\text{ applicable},\\
G_{\mathrm{rules}},&\text{otherwise}.
\end{cases}
\end{equation}
A critical violation therefore determines the score, while unavailable independent evidence produces \(\bot\) rather than implicit credit.

\begin{algorithm}[t]
\caption{Trajectory tracing of harness intervention dynamics}
\label{alg:tracing}
\begin{algorithmic}[1]

\Require
Agent \(A\), task set \(\mathcal{X}\), environment \(\mathcal{E}\),
harness pair \((H_{\mathrm{nat}},H_{\mathrm{med}})\),
repetitions \(R\)

\Ensure
Mechanism evidence \(\textsc{Mechanism}\);
benefit evidence \(\textsc{Benefit}\)

\State
\(\mathcal{H}_{\Delta}
\leftarrow
\mathcal{H}(H_{\mathrm{med}})
\setminus
\mathcal{H}(H_{\mathrm{nat}})\)

\ForAll{\(x\in\mathcal{X},\, k=1,\dots,R\)}

    \State
    \(\tau_{\mathrm{nat}}
    \leftarrow
    \operatorname{Run}(A,H_{\mathrm{nat}},\mathcal{E},x)\)

    \State
    \(\tau_{\mathrm{med}}
    \leftarrow
    \operatorname{Run}(A,H_{\mathrm{med}},\mathcal{E},x)\)

    \ForAll{\(m\in\mathcal{H}_{\Delta}\) eligible in \(\tau_{\mathrm{med}}\)}

        \State
        \(\pi_{x,k}(m)
        \leftarrow
        \max\{z_i^{m}: z_i^{m}\text{ is evidenced in }\tau_{\mathrm{med}}\}\)

    \EndFor

    \State
    \(\Pi_{x,k}
    \leftarrow
    \{m:\pi_{x,k}(m)=z_5^{m}\}\)

    \State
    \(\Delta\mathbf{S}_{x,k}
    \leftarrow
    \mathbf{S}(\tau_{\mathrm{med}})
    -
    \mathbf{S}(\tau_{\mathrm{nat}})\)

    \State
    \(\Delta Y_{x,k}
    \leftarrow
    Y(\tau_{\mathrm{med}})
    -
    Y(\tau_{\mathrm{nat}})\)

\EndFor

\For{\(j=1,\dots,5\)}

    \State
    \(\displaystyle
    \rho_j
    \leftarrow
    \frac{
    \#\{(m,x,k):\pi_{x,k}(m)\succeq z_j\}
    }{
    \#\{(m,x,k):\pi_{x,k}(m)\succeq z_{j-1}\}
    }\)

\EndFor

\State
\(\textsc{Mechanism}
\leftarrow
(\{\rho_j\},
\Pr[\Pi_{x,k}\neq\emptyset])\)

\State
\(\textsc{Benefit}
\leftarrow
\operatorname{PairedTest}
(\{\Delta\mathbf{S}_{x,k}\},
\{\Delta Y_{x,k}\})\)

\State
\Return
\((\textsc{Mechanism},\textsc{Benefit})\)

\end{algorithmic}
\end{algorithm}

Indicators excluded during selection remain available for diagnostic analysis but do not contribute to the reported process scores. For example, evidence precision may reward retrieving fewer facts, generic checking events may not establish closed-loop verification, and explicit uptake may be unnecessary when an agent directly uses structured evidence. Native Outcome and the seven process dimensions are therefore reported separately.
\subsubsection{Trajectory Tracing of Harness Intervention Dynamics}
\label{sec:trace}

Efficiency and Safety localize process quality, but do not show whether a MedHarness intervention actually reaches and influences the agent. We therefore represent each intervention \(m\) as a trajectory-level propagation process:

\begin{equation}
\label{eq:intervention-state}
z_0^{m}
\rightarrow
z_1^{m}
\rightarrow
z_2^{m}
\rightarrow
z_3^{m}
\rightarrow
z_4^{m}
\rightarrow
z_5^{m}.
\end{equation}

where \(z_i^{m}\) denotes the \(i\)-th state: eligible, fired, control-acquired, delivered, uptake, and behavior. Each transition provides stronger downstream evidence that the intervention has propagated through the execution loop. Checkpoints and terminal outcomes remain external evaluation signals and are quantified through paired comparisons rather than treated as propagation states. Algorithm~\ref{alg:tracing} formalizes this procedure.

\section{Experimental Setup}
\label{sec:experiments}

\paragraph{Models and configurations.}
We evaluate eight API-based agents from seven providers: GPT-5.5~\cite{openai2026gpt55} and GPT-5.4~\cite{openai2026gpt54} from OpenAI, Qwen3.7-Max~\cite{qwen2026qwen37} from Qwen, Claude Sonnet 5~\cite{anthropic2026sonnet5} from Anthropic, Kimi K2.7 Code~\cite{moonshot2026kimi27} from Moonshot AI, DeepSeek-V4-Flash~\cite{deepseek2026v4} from DeepSeek AI, MiniMax-M3~\cite{minimax2026m3} from MiniMax, and Gemini 3 Flash~\cite{googledeepmind2025gemini3flash} from Google DeepMind. All models are evaluated without parameter updates or task-specific fine-tuning. Following Section~\ref{sec:method}, we compare the benchmark-native runtime \(H_{\mathrm{nat}}\) with the healthcare-specific MedHarness runtime \(H_{\mathrm{med}}\). Within each pair, the task, model, system prompt, tool inventory, transport, and execution budgets are fixed. Each configuration is run three times, with every run beginning from a fresh conversation and a restored environment snapshot. PhysicianBench and HealthAdminBench permit at most 100 environment actions per task, MedCTA permits 10, and each model turn is capped at 16,000 output tokens.

\paragraph{Metrics.}
Following PhysicianBench~\cite{liu2026physicianbench}, we estimate the empirical single-run success probability \(p_x\) from three independent runs for each task \(x\) and compute \(\mathrm{Pass@3}_x=1-(1-p_x)^3\). The reported Pass@3 is macro-averaged across tasks. Subtask performance and each process score in F/R/E/W/A/V/B are averaged across the three runs. Process aggregates include only trajectories with applicable and sufficient evidence, and their evaluation support is reported separately. For the controlled perturbation study, we report target sensitivity, selectivity, balanced accuracy, macro-F1, permutation significance, and bootstrap confidence intervals.
\section{Results and Analysis}
\label{sec:results}

\subsection{MedHarness across Agents and Healthcare Environments}
\label{sec:static-results}

The first experiment examines whether healthcare-specific runtime support improves agent execution across models and environments. We compare Native and MedHarness across eight agents and three healthcare environments, measuring terminal success, intermediate completion, and process quality under otherwise fixed conditions. As shown in Table~\ref{tab:static-results}, MedHarness improves Pass@3 in 19 of 24 agent--environment configurations and Subtask performance in 22 of 24. Patient--Evidence Alignment improves in all 24 configurations, Closed-Loop Verification in 23, and Care Action Fulfillment in 22. The gains nevertheless differ across evaluation levels. For GPT-5.4 on HealthAdminBench, Subtask increases from 0.543 to 0.633 while Pass@3 decreases from 0.267 to 0.105; on PhysicianBench, Pass@3 increases from 0.330 to 0.360 while Subtask decreases from 0.777 to 0.732. This divergence shows that terminal, intermediate, and process performance capture complementary effects of MedHarness.

\begin{table}[htbp]
\caption{
Native and MedHarness performance across eight agents and three healthcare environments.
Pass@3 measures terminal task success, Subtask measures intermediate task completion, and F/R/E/W/A/V/B denote the seven healthcare-specific process dimensions.
Values in parentheses indicate changes relative to Native.
Bold denotes the best score within each environment and runtime configuration.
}
\label{tab:static-results}
\centering
\tiny
\setlength{\tabcolsep}{1.6pt}
\renewcommand{\arraystretch}{1.05}

\resizebox{\textwidth}{!}{%
\begin{tabular}{
    l
    S[table-format=1.3]
    S[table-format=1.3]
    S[table-format=1.3]
    S[table-format=1.3]
    S[table-format=1.3]
    S[table-format=1.3]
    S[table-format=1.3]
    S[table-format=1.3]
    S[table-format=1.3]
    @{\hspace{2.5pt}\vrule width 0.45pt\hspace{2.5pt}}
    c c c c c c c c c
}
\toprule

& \multicolumn{9}{c}{\textbf{Native}}
& \multicolumn{9}{c}{\textbf{Medical Harness}} \\

\cmidrule(lr){2-10}
\cmidrule(lr){11-19}

\textbf{Model}
& {\textbf{Pass@3}}
& {\textbf{Subtask}}
& {\textbf{F}}
& {\textbf{R}}
& {\textbf{E}}
& {\textbf{W}}
& {\textbf{A}}
& {\textbf{V}}
& {\textbf{B}}
& {\textbf{Pass@3}}
& {\textbf{Subtask}}
& {\textbf{F}}
& {\textbf{R}}
& {\textbf{E}}
& {\textbf{W}}
& {\textbf{A}}
& {\textbf{V}}
& {\textbf{B}} \\

\midrule


\rowcolor{benchadmin}
\multicolumn{19}{c}{\textbf{HealthAdminBench} (\(n=135\))} \\

Claude Sonnet 5
& \best{0.341} & \best{0.794}
& 0.904 & 0.944 & 0.602 & 0.941 & 0.910 & 0.351 & 0.568
& \dcell{0.363}{\gain{0.022}}
& \dcell{\best{0.805}}{\gain{0.011}}
& \dcell{0.917}{\gain{0.013}}
& \dcell{0.946}{\gain{0.002}}
& \dcell{0.622}{\gain{0.020}}
& \dcell{0.942}{\gain{0.001}}
& \dcell{0.912}{\gain{0.002}}
& \dcell{0.649}{\gain{0.298}}
& \dcell{0.615}{\gain{0.047}} \\

GPT-5.4
& 0.267 & 0.543
& 0.648 & 0.696 & 0.522 & 0.824 & 0.916 & 0.283 & \best{0.774}
& \dcell{0.105}{\loss{0.162}}
& \dcell{0.633}{\gain{0.090}}
& \dcell{0.868}{\gain{0.220}}
& \dcell{0.913}{\gain{0.217}}
& \dcell{0.604}{\gain{0.082}}
& \dcell{0.902}{\gain{0.078}}
& \dcell{0.906}{\loss{0.010}}
& \dcell{0.589}{\gain{0.306}}
& \dcell{0.670}{\loss{0.104}} \\

Qwen3.7-Max
& 0.252 & 0.762
& 0.907 & 0.933 & 0.632 & 0.898 & \best{0.920} & 0.375 & 0.709
& \dcell{\best{0.481}}{\gain{0.229}}
& \dcell{0.786}{\gain{0.024}}
& \dcell{\best{0.944}}{\gain{0.037}}
& \dcell{0.934}{\gain{0.001}}
& \dcell{0.637}{\gain{0.005}}
& \dcell{0.896}{\loss{0.002}}
& \dcell{0.906}{\loss{0.014}}
& \dcell{0.677}{\gain{0.302}}
& \dcell{0.739}{\gain{0.030}} \\

GPT-5.5
& 0.244 & 0.760
& \best{0.922} & 0.949 & \best{0.727} & \best{0.962}
& 0.909 & \best{0.389} & 0.741
& \dcell{0.267}{\gain{0.023}}
& \dcell{0.792}{\gain{0.032}}
& \dcell{0.937}{\gain{0.015}}
& \dcell{\best{0.963}}{\gain{0.014}}
& \dcell{\best{0.821}}{\gain{0.094}}
& \dcell{\best{0.964}}{\gain{0.002}}
& \dcell{0.912}{\gain{0.003}}
& \dcell{\best{0.701}}{\gain{0.312}}
& \dcell{\best{0.756}}{\gain{0.015}} \\

Kimi K2.7 Code
& 0.244 & 0.647
& 0.900 & 0.947 & 0.643 & 0.935 & 0.919 & 0.367 & 0.715
& \dcell{0.274}{\gain{0.030}}
& \dcell{0.726}{\gain{0.079}}
& \dcell{0.931}{\gain{0.031}}
& \dcell{0.954}{\gain{0.007}}
& \dcell{0.645}{\gain{0.002}}
& \dcell{0.938}{\gain{0.003}}
& \dcell{0.915}{\loss{0.004}}
& \dcell{0.677}{\gain{0.310}}
& \dcell{0.634}{\loss{0.081}} \\

MiniMax-M3
& 0.222 & 0.685
& 0.911 & 0.928 & 0.658 & 0.943 & 0.870 & 0.380 & 0.681
& \dcell{0.289}{\gain{0.067}}
& \dcell{0.697}{\gain{0.012}}
& \dcell{0.919}{\gain{0.008}}
& \dcell{0.918}{\loss{0.010}}
& \dcell{0.693}{\gain{0.035}}
& \dcell{0.947}{\gain{0.004}}
& \dcell{0.884}{\gain{0.014}}
& \dcell{0.628}{\gain{0.248}}
& \dcell{0.630}{\loss{0.051}} \\

DeepSeek-V4-Flash
& 0.193 & 0.570
& 0.911 & \best{0.950} & 0.609 & 0.950 & 0.907 & 0.341 & 0.693
& \dcell{0.178}{\loss{0.015}}
& \dcell{0.577}{\gain{0.007}}
& \dcell{0.912}{\gain{0.001}}
& \dcell{0.887}{\loss{0.063}}
& \dcell{0.628}{\gain{0.019}}
& \dcell{0.881}{\loss{0.069}}
& \dcell{\best{0.936}}{\gain{0.029}}
& \dcell{0.572}{\gain{0.231}}
& \dcell{0.707}{\gain{0.014}} \\

Gemini 3 Flash
& 0.156 & 0.700
& 0.893 & 0.940 & 0.673 & 0.935 & 0.908 & 0.371 & 0.681
& \dcell{0.267}{\gain{0.111}}
& \dcell{0.715}{\gain{0.015}}
& \dcell{0.898}{\gain{0.005}}
& \dcell{0.942}{\gain{0.002}}
& \dcell{0.678}{\gain{0.005}}
& \dcell{0.944}{\gain{0.009}}
& \dcell{0.913}{\gain{0.005}}
& \dcell{0.625}{\gain{0.254}}
& \dcell{0.701}{\gain{0.020}} \\

\midrule


\rowcolor{benchphys}
\multicolumn{19}{c}{\textbf{PhysicianBench} (\(n=100\))} \\

GPT-5.5
& \best{0.380} & \best{0.801}
& \best{0.872} & 0.942 & 0.776 & 0.910 & 0.973 & \best{0.464} & 0.511
& \dcell{\best{0.450}}{\gain{0.070}}
& \dcell{\best{0.818}}{\gain{0.017}}
& \dcell{0.877}{\gain{0.005}}
& \dcell{0.937}{\loss{0.005}}
& \dcell{0.784}{\gain{0.008}}
& \dcell{0.912}{\gain{0.002}}
& \dcell{0.966}{\loss{0.007}}
& \dcell{0.496}{\gain{0.032}}
& \dcell{0.512}{\gain{0.001}} \\

GPT-5.4
& 0.330 & 0.777
& 0.860 & 0.940 & 0.774 & 0.912 & \best{0.975} & 0.457 & \best{0.621}
& \dcell{0.360}{\gain{0.030}}
& \dcell{0.732}{\loss{0.045}}
& \dcell{0.888}{\gain{0.028}}
& \dcell{0.918}{\loss{0.022}}
& \dcell{0.788}{\gain{0.014}}
& \dcell{\best{0.919}}{\gain{0.007}}
& \dcell{\best{0.980}}{\gain{0.005}}
& \dcell{0.469}{\gain{0.012}}
& \dcell{\best{0.680}}{\gain{0.059}} \\

Qwen3.7-Max
& 0.200 & 0.679
& 0.842 & 0.932 & \best{0.786} & 0.894 & 0.958 & 0.423 & 0.436
& \dcell{0.190}{\loss{0.010}}
& \dcell{0.682}{\gain{0.003}}
& \dcell{0.908}{\gain{0.066}}
& \dcell{0.945}{\gain{0.013}}
& \dcell{\best{0.790}}{\gain{0.004}}
& \dcell{0.910}{\gain{0.016}}
& \dcell{0.957}{\loss{0.001}}
& \dcell{\best{0.522}}{\gain{0.099}}
& \dcell{0.399}{\loss{0.037}} \\

DeepSeek-V4-Flash
& 0.190 & 0.621
& 0.851 & 0.913 & 0.708 & 0.840 & 0.878 & 0.364 & 0.318
& \dcell{0.220}{\gain{0.030}}
& \dcell{0.689}{\gain{0.068}}
& \dcell{\best{0.909}}{\gain{0.058}}
& \dcell{0.933}{\gain{0.020}}
& \dcell{0.772}{\gain{0.064}}
& \dcell{0.852}{\gain{0.012}}
& \dcell{0.882}{\gain{0.004}}
& \dcell{0.462}{\gain{0.098}}
& \dcell{0.486}{\gain{0.168}} \\

MiniMax-M3
& 0.170 & 0.636
& 0.722 & 0.783 & 0.668 & 0.807 & 0.921 & 0.352 & 0.392
& \dcell{0.190}{\gain{0.020}}
& \dcell{0.688}{\gain{0.052}}
& \dcell{0.851}{\gain{0.129}}
& \dcell{0.932}{\gain{0.149}}
& \dcell{0.762}{\gain{0.094}}
& \dcell{0.825}{\gain{0.018}}
& \dcell{0.939}{\gain{0.018}}
& \dcell{0.449}{\gain{0.097}}
& \dcell{0.583}{\gain{0.191}} \\

Claude Sonnet 5
& 0.170 & 0.781
& 0.775 & 0.897 & 0.758 & 0.843 & 0.923 & 0.400 & 0.389
& \dcell{0.220}{\gain{0.050}}
& \dcell{0.787}{\gain{0.006}}
& \dcell{0.798}{\gain{0.023}}
& \dcell{0.909}{\gain{0.012}}
& \dcell{0.762}{\gain{0.004}}
& \dcell{0.851}{\gain{0.008}}
& \dcell{0.927}{\gain{0.004}}
& \dcell{0.494}{\gain{0.094}}
& \dcell{0.441}{\gain{0.052}} \\

Gemini 3 Flash
& 0.070 & 0.557
& 0.870 & \best{0.945} & 0.672 & \best{0.913} & 0.959 & 0.342 & 0.395
& \dcell{0.090}{\gain{0.020}}
& \dcell{0.552}{\loss{0.005}}
& \dcell{0.883}{\gain{0.013}}
& \dcell{\best{0.958}}{\gain{0.013}}
& \dcell{0.696}{\gain{0.024}}
& \dcell{0.917}{\gain{0.004}}
& \dcell{0.959}{\scriptsize\((0.000)\)}
& \dcell{0.491}{\gain{0.149}}
& \dcell{0.438}{\gain{0.043}} \\

Kimi K2.7 Code
& 0.020 & 0.319
& 0.343 & 0.504 & 0.278 & 0.668 & 0.961 & 0.143 & 0.491
& \dcell{0.150}{\gain{0.130}}
& \dcell{0.594}{\gain{0.275}}
& \dcell{0.756}{\gain{0.413}}
& \dcell{0.868}{\gain{0.364}}
& \dcell{0.704}{\gain{0.426}}
& \dcell{0.786}{\gain{0.118}}
& \dcell{0.867}{\loss{0.094}}
& \dcell{0.406}{\gain{0.263}}
& \dcell{0.498}{\gain{0.007}} \\

\midrule


\rowcolor{benchcta}
\multicolumn{19}{c}{\textbf{MedCTA} (\(n=107\))} \\

Qwen3.7-Max
& \best{0.729} & 0.555
& 0.693 & 0.775 & 0.670 & 0.863 & 0.855 & 0.668 & 0.386
& \dcell{0.720}{\loss{0.009}}
& \dcell{\best{0.581}}{\gain{0.026}}
& \dcell{0.681}{\loss{0.012}}
& \dcell{\best{0.836}}{\gain{0.061}}
& \dcell{0.679}{\gain{0.009}}
& \dcell{\best{0.873}}{\gain{0.010}}
& \dcell{0.868}{\gain{0.013}}
& \dcell{0.676}{\gain{0.008}}
& \dcell{0.579}{\gain{0.193}} \\

GPT-5.5
& 0.710 & 0.534
& 0.665 & 0.762 & 0.660 & 0.839 & 0.874 & 0.657 & 0.488
& \dcell{0.687}{\loss{0.023}}
& \dcell{0.556}{\gain{0.022}}
& \dcell{0.673}{\gain{0.008}}
& \dcell{0.769}{\gain{0.007}}
& \dcell{0.675}{\gain{0.015}}
& \dcell{0.838}{\loss{0.001}}
& \dcell{0.884}{\gain{0.010}}
& \dcell{0.604}{\loss{0.053}}
& \dcell{\best{0.709}}{\gain{0.221}} \\

Claude Sonnet 5
& 0.701 & 0.475
& 0.650 & 0.728 & 0.592 & 0.840 & 0.854 & 0.593 & 0.448
& \dcell{\best{0.758}}{\gain{0.057}}
& \dcell{0.481}{\gain{0.006}}
& \dcell{0.696}{\gain{0.046}}
& \dcell{0.764}{\gain{0.036}}
& \dcell{0.671}{\gain{0.079}}
& \dcell{0.801}{\loss{0.039}}
& \dcell{0.874}{\gain{0.020}}
& \dcell{0.683}{\gain{0.090}}
& \dcell{0.555}{\gain{0.107}} \\

GPT-5.4
& 0.682 & 0.516
& \best{0.706} & \best{0.792} & \best{0.709} & \best{0.865}
& 0.870 & \best{0.705} & 0.436
& \dcell{0.745}{\gain{0.063}}
& \dcell{0.574}{\gain{0.058}}
& \dcell{\best{0.719}}{\gain{0.013}}
& \dcell{0.799}{\gain{0.007}}
& \dcell{\best{0.720}}{\gain{0.011}}
& \dcell{0.872}{\gain{0.007}}
& \dcell{0.864}{\loss{0.006}}
& \dcell{\best{0.710}}{\gain{0.005}}
& \dcell{0.638}{\gain{0.202}} \\

DeepSeek-V4-Flash
& 0.645 & 0.536
& 0.591 & 0.705 & 0.531 & 0.795 & 0.861 & 0.537 & 0.561
& \dcell{0.664}{\gain{0.019}}
& \dcell{0.561}{\gain{0.025}}
& \dcell{0.618}{\gain{0.027}}
& \dcell{0.740}{\gain{0.035}}
& \dcell{0.607}{\gain{0.076}}
& \dcell{0.817}{\gain{0.022}}
& \dcell{0.879}{\gain{0.018}}
& \dcell{0.623}{\gain{0.086}}
& \dcell{0.614}{\gain{0.053}} \\

Kimi K2.7 Code
& 0.645 & \best{0.569}
& 0.612 & 0.735 & 0.597 & 0.812 & 0.869 & 0.598 & 0.642
& \dcell{0.686}{\gain{0.041}}
& \dcell{0.570}{\gain{0.001}}
& \dcell{0.614}{\gain{0.002}}
& \dcell{0.743}{\gain{0.008}}
& \dcell{0.602}{\gain{0.005}}
& \dcell{0.815}{\gain{0.003}}
& \dcell{0.875}{\gain{0.006}}
& \dcell{0.620}{\gain{0.022}}
& \dcell{0.688}{\gain{0.046}} \\

MiniMax-M3
& 0.570 & 0.523
& 0.606 & 0.712 & 0.573 & 0.804 & 0.859 & 0.588 & 0.494
& \dcell{0.664}{\gain{0.094}}
& \dcell{0.542}{\gain{0.019}}
& \dcell{0.580}{\loss{0.026}}
& \dcell{0.748}{\gain{0.036}}
& \dcell{0.603}{\gain{0.030}}
& \dcell{0.790}{\loss{0.014}}
& \dcell{0.884}{\gain{0.025}}
& \dcell{0.589}{\gain{0.001}}
& \dcell{\best{0.709}}{\gain{0.215}} \\

Gemini 3 Flash
& 0.561 & 0.530
& 0.496 & 0.725 & 0.525 & 0.742 & \best{0.904} & 0.531 & \best{0.721}
& \dcell{0.738}{\gain{0.177}}
& \dcell{0.548}{\gain{0.018}}
& \dcell{0.587}{\gain{0.091}}
& \dcell{0.815}{\gain{0.090}}
& \dcell{0.640}{\gain{0.115}}
& \dcell{0.810}{\gain{0.068}}
& \dcell{\best{0.885}}{\loss{0.019}}
& \dcell{0.565}{\gain{0.034}}
& \dcell{0.627}{\loss{0.094}} \\

\bottomrule
\end{tabular}%
}
\end{table}

\subsection{Comparison with General-Purpose Harnesses}
\label{sec:static-comparison}

The second experiment tests whether healthcare-specific harness design provides benefits beyond general-purpose runtime support. Using GPT-5.5 as the target agent, we compare MedHarness with Claude Code, NeMo Guardrails, and OpenAI Agents SDK on 60 tasks, obtained by stratifying the native task categories of each benchmark and sampling 20 tasks per environment. Table~\ref{tab:static-harness-comparison} shows that MedHarness improves Pass@3 and Subtask over Native on all three benchmarks and achieves the highest average process score in each environment. The general-purpose harnesses improve only three of nine Pass@3 comparisons and two of nine Subtask comparisons, indicating that isolated process gains do not consistently translate into healthcare task completion.

\begin{table}[htbp]
\caption{
Comparison of MedHarness with general-purpose harnesses using GPT-5.5 as the target agent.
For each healthcare environment, 20 tasks are selected by stratified sampling over native task categories.
Pass@3 measures terminal task success, Subtask measures intermediate task completion, and F/R/E/W/A/V/B denote the seven healthcare-specific process dimensions.
Avg. is the mean of the seven process scores.
Bold denotes the best score within each environment.
}
\label{tab:static-harness-comparison}
\centering
\scriptsize
\setlength{\tabcolsep}{3.2pt}

\begin{tabular}{
l
S[table-format=1.3]
S[table-format=1.3]
S[table-format=1.3]
S[table-format=1.3]
S[table-format=1.3]
S[table-format=1.3]
S[table-format=1.3]
S[table-format=1.3]
S[table-format=1.3]
S[table-format=1.3]
}
\toprule
\textbf{Method}
& {\textbf{Pass@3}}
& {\textbf{Subtask}}
& {\textbf{F}}
& {\textbf{R}}
& {\textbf{E}}
& {\textbf{W}}
& {\textbf{A}}
& {\textbf{V}}
& {\textbf{B}}
& {\textbf{Avg.}} \\
\midrule

\rowcolor{tablegray}
\multicolumn{11}{l}{\textbf{PhysicianBench} ($n=20$)} \\

Native
& 0.310 & 0.677
& 0.833 & 0.934 & 0.740 & 0.884 & 0.966 & 0.337 & 0.529
& 0.746 \\

MedHarness
& \bfseries 0.720 & \bfseries 0.963
& \bfseries 0.908 & \bfseries 0.946 & \bfseries 0.800
& \bfseries 0.921 & \bfseries 0.968 & \bfseries 0.390 & 0.638
& \bfseries 0.796 \\

Claude Code
& 0.280 & 0.322
& 0.558 & 0.769 & 0.650 & 0.804 & 0.961 & 0.317 & \bfseries 0.748
& 0.687 \\

NeMo Guardrails
& 0.430 & 0.185
& 0.442 & 0.744 & 0.630 & 0.709 & 0.931 & 0.300 & 0.435
& 0.599 \\

OpenAI Agents SDK
& 0.250 & 0.192
& 0.392 & 0.731 & 0.550 & 0.670 & 0.938 & 0.296 & 0.513
& 0.584 \\

\midrule

\rowcolor{tablegray}
\multicolumn{11}{l}{\textbf{MedCTA} ($n=20$)} \\

Native
& 0.669 & 0.500
& 0.692 & 0.760 & \bfseries 0.654
& \bfseries 0.845 & 0.864 & 0.693 & 0.297
& 0.686 \\

MedHarness
& \bfseries 0.846 & \bfseries 0.679
& \bfseries 0.729 & \bfseries 0.773 & \bfseries 0.654
& 0.814 & 0.866 & \bfseries 0.701 & 0.763
& \bfseries 0.757 \\

Claude Code
& 0.167 & 0.438
& 0.154 & 0.446 & 0.269 & 0.577 & 0.864 & 0.282 & 0.749
& 0.477 \\

NeMo Guardrails
& 0.692 & 0.564
& 0.538 & 0.703 & 0.577 & 0.753 & 0.885 & 0.603 & 0.806
& 0.695 \\

OpenAI Agents SDK
& 0.538 & 0.538
& 0.577 & 0.738 & 0.577 & 0.782 & \bfseries 0.894 & 0.680 & \bfseries 0.819
& 0.724 \\

\midrule

\rowcolor{tablegray}
\multicolumn{11}{l}{\textbf{HealthAdminBench} ($n=20$)} \\

Native
& 0.500 & 0.566
& 0.629 & 0.988 & \bfseries 0.684
& \bfseries 0.875 & 0.917 & 0.414 & 0.603
& 0.730 \\

MedHarness
& \bfseries 0.680 & \bfseries 0.731
& \bfseries 0.752 & \bfseries 0.993 & 0.590
& 0.794 & 0.933 & \bfseries 0.660 & 0.642
& \bfseries 0.766 \\

Claude Code
& 0.550 & 0.398
& 0.175 & 0.181 & 0.123 & 0.395 & \bfseries 0.978 & 0.096 & 0.614
& 0.366 \\

NeMo Guardrails
& 0.200 & 0.272
& 0.600 & 0.734 & 0.436 & 0.542 & 0.969 & 0.279 & 0.572
& 0.590 \\

OpenAI Agents SDK
& 0.358 & 0.416
& 0.471 & 0.025 & 0.381 & 0.125 & 0.972 & 0.429 & \bfseries 0.656
& 0.437 \\

\bottomrule
\end{tabular}
\end{table}

\subsection{Process Failure Sensitivity and Discrimination}
\label{sec:process-validation}

The third experiment evaluates whether the seven process dimensions are sensitive to their intended failures and sufficiently discriminative for fault localization. We introduce one controlled failure into each applicable clean trajectory while preserving the terminal answer and unrelated evidence. Table~\ref{tab:fault-injection} summarizes the seven intervention families.

\begin{table}[htbp]
\caption{Single-fault interventions used to evaluate the seven process dimensions. F/R/E/W denote the four Efficiency dimensions, while A/V/B denote the three Safety dimensions. Each intervention is adapted to the benchmark environment while preserving the same failure semantics.}
\label{tab:fault-injection}
\centering
\scriptsize
\setlength{\tabcolsep}{5pt}

\begin{tabular}{
p{0.05\textwidth}
p{0.22\textwidth}
p{0.65\textwidth}
}
\toprule
\textbf{Dim.}
& \textbf{Injected failure}
& \textbf{Example transformation} \\
\midrule

F & Incomplete care action
& Remove evidence that an applicable care action or required artifact was completed while preserving the surrounding execution trace. \\

R & Resource-use failure
& Corrupt the selected clinical resource, authorization state, parameter semantics, or use of the returned result while leaving the intended task unchanged. \\

E & Evidence misalignment
& Remove required evidence or alter its subject binding such that the evidence no longer supports the current patient, state, or decision. \\

W & Workflow discontinuity
& Remove evidence required to maintain valid progression, recovery, readiness, or termination across successive workflow states. \\

A & Missing audit evidence
& Remove state, failure, recovery, provenance, or reliability evidence required to reconstruct the execution and its consequential decisions. \\

V & Missing closed-loop verification
& Remove an applicable precondition confirmation or post-action verification while preserving the corresponding action itself. \\

B & Conduct-boundary violation
& Insert an unauthorized, out-of-scope, incorrectly sourced, or otherwise boundary-violating action while preserving the surrounding workflow. \\

\bottomrule
\end{tabular}
\end{table}

\begin{figure*}[tbp]
\centering

\begin{minipage}[t]{0.485\textwidth}
\centering
\includegraphics[width=\linewidth]{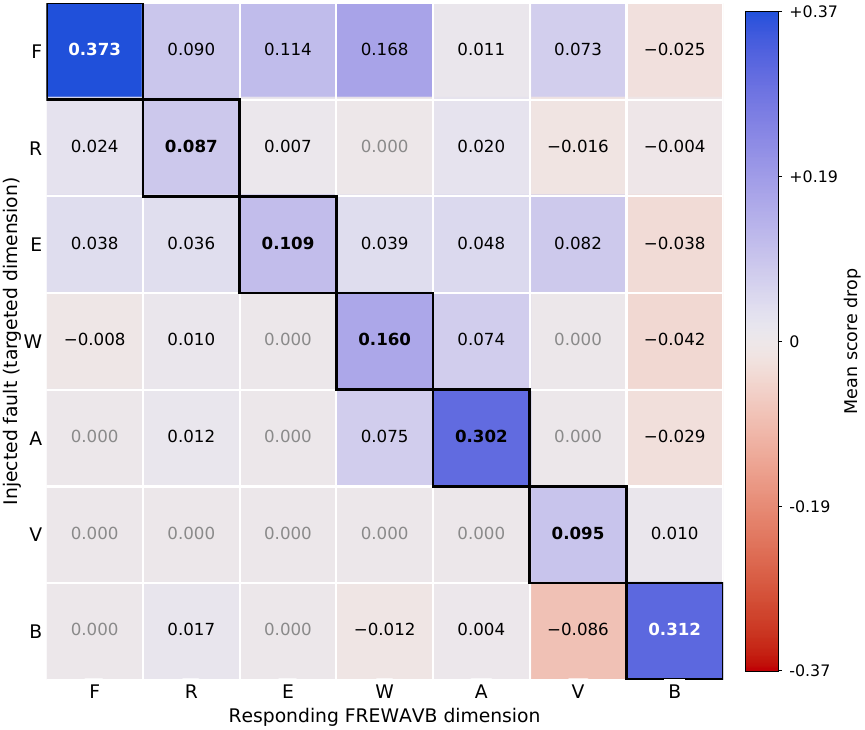}
\vspace{-2mm}
\textbf{(a) Perturbation response matrix}
\end{minipage}
\hfill
\begin{minipage}[t]{0.485\textwidth}
\centering
\includegraphics[width=\linewidth]{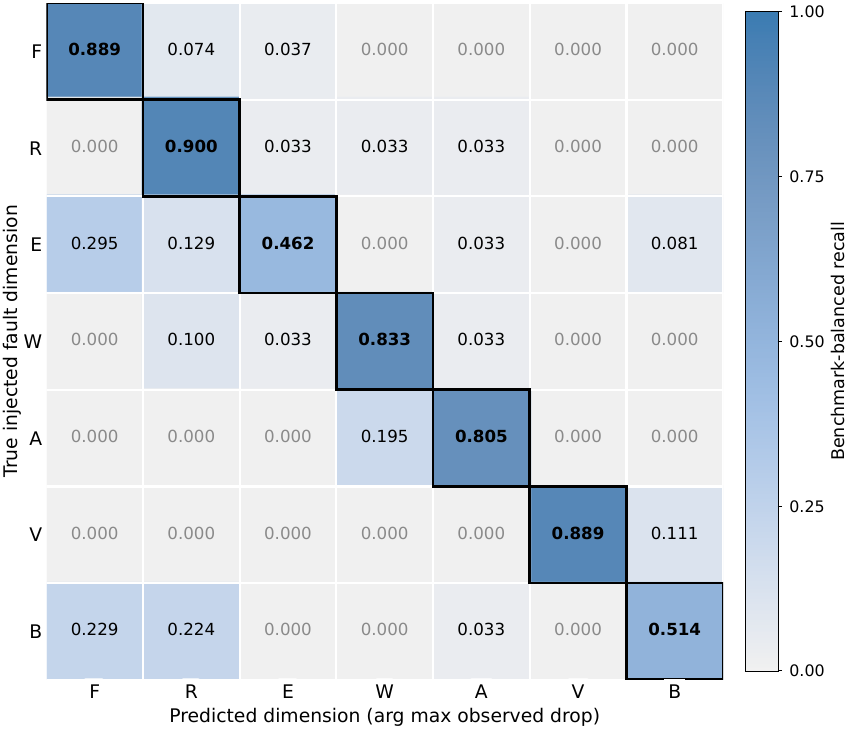}
\vspace{-2mm}
\textbf{(b) Fault-localization confusion matrix}
\end{minipage}

\caption{
Process-failure sensitivity and discrimination under controlled single-fault perturbations.
\textbf{(a)} Benchmark-balanced mean score decrease for each injected failure and responding dimension, computed over the applicable support of the responding dimension.
The targeted dimension is the largest response in all seven rows.
\textbf{(b)} Benchmark-balanced fault-localization confusion matrix obtained by predicting the injected failure from the dimension with the largest observed score decrease.
Rows are normalized by injected failure class.
}
\label{fig:process-validation}
\end{figure*}

For each clean--perturbed pair, we compute the decrease in each applicable process score. Because process dimensions have trajectory-level applicability as defined in Section~\ref{sec:method}, these decreases need not share identical support across dimensions: a responding dimension contributes to an aggregate only when both the clean and perturbed trajectories provide sufficient evidence for that dimension. Figure~\ref{fig:process-validation}(a) therefore reports, for each injected fault and responding dimension, the benchmark-balanced mean decrease over that responding dimension's applicable support.

Selectivity is defined at the pair level rather than from the aggregated response matrix. For a clean--perturbed pair with targeted dimension \(d^\star\), the selectivity margin is
\[
m
=
\Delta S_{d^\star}
-
\max_{d\neq d^\star,\; d\in\mathcal{C}}
\Delta S_d,
\]
where \(\mathcal{C}\) denotes the non-target dimensions that are jointly comparable for that pair. These pair-level margins are then aggregated across benchmarks. Consequently, the selectivity values in Table~\ref{tab:process-sensitivity} are not obtained by subtracting the largest off-diagonal mean in Figure~\ref{fig:process-validation}(a) from the diagonal mean, since the corresponding quantities can be computed over different effective supports.

Sensitivity is determined by the response of the targeted dimension, while discrimination is evaluated by predicting the injected failure from the dimension with the largest observed decrease. Figure~\ref{fig:process-validation} and Table~\ref{tab:process-sensitivity} show that the targeted dimension produces the largest mean decrease for all seven fault classes, with every target-drop and selectivity-margin confidence interval excluding zero. Fault localization reaches a benchmark-balanced accuracy of 0.756 and a macro-F1 of 0.706, significantly above permuted labels (\(p=0.001\)). Patient--Evidence Alignment and Clinical Conduct Boundaries are the most confusable dimensions, with recalls of 0.462 and 0.514, respectively.

\begin{table}[htbp]
\caption{
Sensitivity and selectivity under controlled single-fault perturbations.
Target drop denotes the benchmark-balanced decrease in the targeted process dimension.
Selectivity margin is computed for each clean--perturbed pair as the targeted-dimension drop minus the largest drop among jointly comparable non-target dimensions, and is then aggregated across benchmarks.
Brackets report 95\% bootstrap confidence intervals.
}
\label{tab:process-sensitivity}
\centering
\scriptsize
\setlength{\tabcolsep}{3.8pt}

\begin{tabular}{lcc}
\toprule
\textbf{Dimension}
& \shortstack{\textbf{Target drop}\\\textbf{(95\% CI)}}
& \shortstack{\textbf{Selectivity margin}\\\textbf{(95\% CI)}} \\
\midrule

Care Action Fulfillment (F)
& 0.373 [0.302, 0.440]
& 0.326 [0.258, 0.387] \\

Clinical Resource Competence (R)
& 0.087 [0.071, 0.104]
& 0.086 [0.066, 0.105] \\

Patient--Evidence Alignment (E)
& 0.109 [0.067, 0.155]
& 0.087 [0.042, 0.143] \\

Clinical Workflow Continuity (W)
& 0.160 [0.132, 0.186]
& 0.168 [0.134, 0.201] \\

Clinical Auditability (A)
& 0.302 [0.262, 0.340]
& 0.285 [0.243, 0.324] \\

Closed-Loop Verification (V)
& 0.095 [0.090, 0.101]
& 0.092 [0.084, 0.100] \\

Clinical Conduct Boundaries (B)
& 0.312 [0.195, 0.435]
& 0.328 [0.215, 0.448] \\

\bottomrule
\end{tabular}
\end{table}

\section{Discussion}
\label{sec:discussion}

The results demonstrate that the Medical Harness provides an effective healthcare-specific runtime layer for coordinating evidence, tools, workflow state, verification, and authorization around medical agents. MedHarness improves Outcome in 19 of 24 agent--environment configurations and Subtask performance in 22 of 24, although its effects vary across agents, environments, and process dimensions. This variation reflects differences in how agents use tools, maintain state, respond to runtime support, and propagate interventions into subsequent actions, together with the distinct demands of FHIR/EHR operations, clinical GUI workflows, and multimodal tool use. The Medical Harness Protocol further reveals where these effects arise by evaluating Care Action Fulfillment, Clinical Resource Competence, Patient--Evidence Alignment, Clinical Workflow Continuity, Clinical Auditability, Closed-Loop Verification, and Clinical Conduct Boundaries. These dimensions expose both intermediate improvements that do not lead to terminal success and successful outcomes that conceal incomplete, unverified, or boundary-violating behavior. Nevertheless, MedHarnessBench contains 342 tasks from three controlled research environments and does not represent the full diversity of hospital workflows, tool schemas, institutional policies, or clinical contingencies. Some process dimensions also depend on automated or LLM-based evaluators, and individual harness mechanisms may influence multiple dimensions. The current findings therefore validate the Medical Harness as a measurable research framework but do not establish clinical deployment readiness, regulatory compliance, or patient-safety effectiveness in real-world care.
\section{Conclusion}

We presented the \textit{Medical Harness}, a healthcare-specific runtime framework that coordinates clinical evidence, heterogeneous tools, workflow state, monitoring, recovery, verification, and authorization around medical agents. To evaluate this framework, we introduced the \protocol, which defines seven healthcare-specific process dimensions and traces how harness interventions are activated, delivered, adopted, and propagated through clinical workflows. We further instantiated the protocol as \benchmark, comprising 342 executable tasks across FHIR/EHR operations, healthcare-administration interfaces, and multimodal clinical tools. Across eight agents, MedHarness improves both terminal task performance and intermediate execution, while comparisons with general-purpose harnesses and controlled single-fault perturbations demonstrate the value and diagnostic sensitivity of the proposed framework. These results establish the Medical Harness as a measurable runtime layer for identifying and improving execution failures that conventional terminal evaluations cannot expose.

\bibliographystyle{unsrt}
\bibliography{MedHarness_FINAL}

\end{document}